\documentclass{article}

\PassOptionsToPackage{numbers, compress}{natbib}

\usepackage[preprint]{nips_2018}
\usepackage{graphicx}
\usepackage{caption}
\usepackage{subcaption}

\usepackage[utf8]{inputenc} 
\usepackage[T1]{fontenc}    
\usepackage{hyperref}       
\usepackage{url}            
\usepackage{booktabs}       
\usepackage{amsfonts}       
\usepackage{nicefrac}       
\usepackage{microtype}      

\title{Deep Face Quality Assessment}

\author{
  Vishal Agarwal \\
  Department of Electronics and Electrical Engineering\\
  Indian Institute of Technology Guwahati\\
  India \\
  \texttt{vishal.agarwal@iitg.ac.in} \\
}

\begin{document}

\maketitle

\begin{abstract}
  Face image quality is an important factor in facial recognition systems as its verification and recognition accuracy is highly dependent on the quality of image presented. Rejecting low quality images can significantly increase the accuracy of any facial recognition system. In this project, a simple approach is presented to train a deep convolutional neural network to perform end-to-end face image quality assessment. The work is done in 2 stages : First, generation of quality score label and secondly, training a deep convolutional neural network in a supervised manner to predict quality score between 0 and 1. The generation of quality labels is done by comparing the face image with a template of best quality images and then evaluating the normalized score based on the similarity. 
  
\end{abstract}

\section{Introduction}

Human face is a very dynamic biometric system as compared to other biometric systems such as fingerprint which is largely static. The performance of a facial recognition system highly depends upon the quality of the image that it acquires. The utility of a face image to a facial image recognition can be defined by its image quality\cite{rowden}. Low quality images tend to make any facial recognition to perform worse. The various factors that results in false recognition are variations in pose, illumination, occlusion, expression, age, lifestyle, etc.\cite{lit} Under controlled acquisition environment such as uniform lighting, frontal facial pose, neutral expression, high resolution image, etc. the facial recognition system can achieve very low False Acceptance Rate (FAR) in comparison to images taken in the wild\cite{int2, int1}. Before processing the face image for verification or recognition, we can do a quality-check assessment as a pre-processing step. Depending upon the score, the system may decide to reject low quality images and only process certain qualified images for verification or recognition. A critical application for the assessment is Negative Identification Systems such as security checks at banks or airports where suspects try to provide low quality image to evade recognition\cite{rowden}. In such cases, the system should flag such users and access should be provided only after providing perfect aligned facial image.

The primary goal of this project is to develop an end-to-end system for automatic facial quality assessment. Instead of using hand-engineered feature designed approach, a data-driven, transfer learning approach is implemented in this work. The overview of the approach is as follows. First, a database of quality score is generated as similarity score by comparing facial images with a gallery of images using Google's FaceNet embeddings\cite{facenet}. To predict the desired quality score, a pre-trained FaceNet model with Inception v3 architecture and custom added layers is used which outputs value between 0 and 1. This model is trained using the generated quality labelled facial data in a supervised setting.

\section{Previous work}

There are generally two ways that exists in literature to come up with a quality metric : one is Full-Reference based and the other is No-Reference based. Based on the above classification, there are mainly two approaches that is used by most people. First, using some facial textures and properties such as resolution, pose, illumination, etc., design hand-engineered features and functions to predict an absolute quality index\cite{work1}. Secondly, we can compare the image under consideration to a standard reference image and use some comparison technique to get our desired quality metric\cite{work2, work3}. However, the effectiveness of these methods are limited by the applicability of artificially defined facial properties and empirically selected reference image and may not generalize well to different databases or face images with multiple quality factors present.

\begin{figure}[]
\centering
\includegraphics[width=12cm]{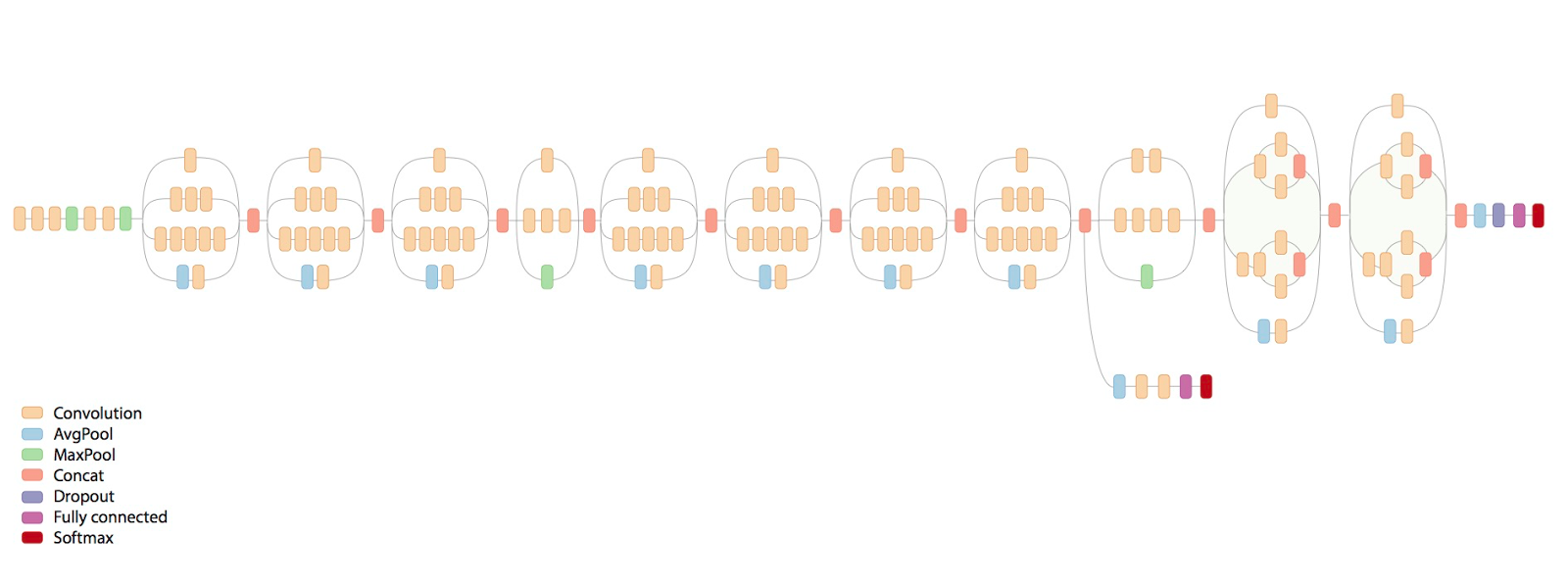}
\caption{Inception v3 architecture}
\label{inception}
\end{figure}


\section{Methodology}

The project can be divided into 2 stages : First, prepare the quality score data by computing the similarity score from a face matching technique. Secondly, using the data generated, end-to-end training of a model which predicts facial image quality score. This trained model will be used for an automatic No Reference face image quality assessment. To get better performance in the task, a transfer learning setup is used by using the pre-trained FaceNet weights\footnote{\url{https://drive.google.com/file/d/1971Xk5RwedbudGgTIrGAL4F7Aifu7id1/view}}. Transfer Learning is a popular method in deep learning where models trained for some task are reused in some closely related tasks\cite{tf}.  Using pre-trained models boost performance as it provides a good starting point and achieve better convergence.

\subsection{Quality score label generation}
The face image data labelled by subjects is divided into two sets consisting of template and probe images for all subjects. The template contains exactly one image per subject which serves as our reference for face similarity matching and should be of best quality. The probe contains rest of the images, except those chosen as template, labelled by the subjects. These serve as our gallery for comparison against template images and generate target quality score. 

The target quality score computed from similarity score from a face matcher is highly correlated with the automatic recognition performance and serves as an 'oracle' for quality measure\cite{rowden}. The FaceNet embedding proposed by Schroff et al.\cite{facenet} has been used to generate face embeddings in an Euclidean space. The simple Euclidean distance between the embeddings gives us the similarity measure between two faces. For each subject, the template image is compared with the probe images by evaluating the comparison score which is the distance between two embeddings. The comparison score is assigned to the probe image under the assumption that the quality of template is as good as a probe image and probe consists of low quality images. As defined in \cite{meth1, meth2, meth3}, the target quality value is a measure of separation between sample's genuine score and its impostor distribution when it is compared with a gallery of template images. Therefore, a normalized comparison score for the jth probe image of ith subject is given by 
\begin{equation} \label{norm}
s_{ij} = \frac{d_{ij}^{G} - \mu_{ij}^{I}}{\sigma_{ij}^{I}}
\end{equation}

\subsection{End-to-End Model}

FaceNet is a simple model which learns mapping from facial image into an Euclidean space. The idea behind the model is faces of same person should lie closer and faces of distinct person should be far away. Based on this principle, they proposed a triplet loss function which helps to learn robust embeddings. The FaceNet model model simplifies facial recognition systems related tasks such as face verification can be done by simply thresholding distance between the embeddings, face recognition can be done by k-Nearest Neighbour and face clustering can be done using k-means clustering.
\begin{equation}
    Triplet \; loss = \sum_{i=1}^{N} \left [ \left \| f(x_i^{a}) - f(x_i^{p}) \right \|_{2}^{2} - \left \| f(x_i^{a}) - f(x_i^{n}) \right \|_{2}^{2} \right ]_+
\end{equation}

The pre-trained FaceNet model based on Inception v3 architecture is used for the task. Since the model outputs a 128-dimensional embedding vector, a 1-dimensional custom layer is added at the end. A sigmoid activation function is used at the last layer so that we get an output value between 0 and 1. The normalized comparison score data serves as a target quality value for training this model. This model is used for end-to-end face quality assessment. The quality score obtained can be used in facial recognition systems to set a threshold according to sensitivity of application and minimal quality requirement for verification. \\

\begin{figure}[]
\centering
\fbox{\includegraphics[width=7cm]{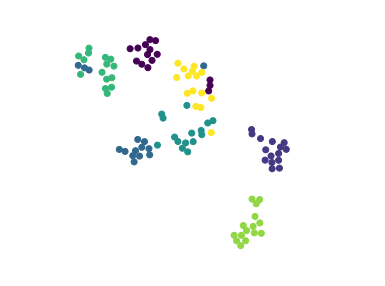}}
\caption{t-SNE of FaceNet embeddings}
\label{tsne}
\end{figure}

\section{Experiment}

\subsection{Dataset}

Labeled Faces in the Wild or LFW database\cite{lfw} has been used for this project. The dataset contains 13233 images of 5749 subjects. There are 1680 subjects with two or more images so one of the images for these 1680 subjects are placed in template gallery and rest 7484 images are used as probe images. The remaining 4069 subjects have just a single image each so these images have been used to extend template gallery. Further, the quality score is generated for 7484 probe images and these are used for training the model in a supervised setting. The model is evaluated on the Georgia Tech Face Database\cite{gtdb} which contains 50 subjects with 15 images per subject.

\begin{figure}[h]
\centering
\includegraphics[width=9cm]{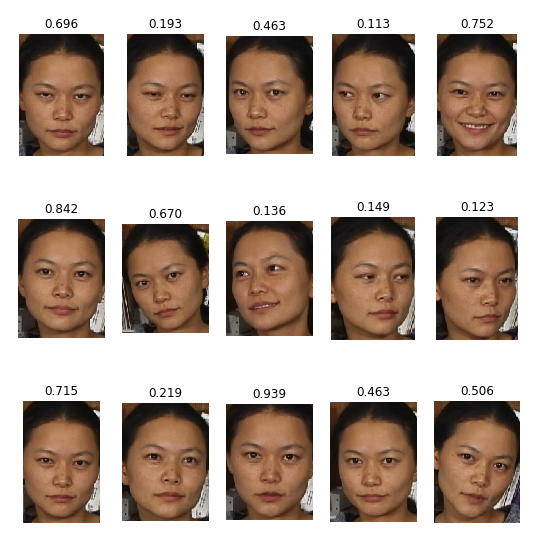}
\caption{Face quality assessment score on a sample subject}
\label{score}
\end{figure}

\subsection{Setup}

The model takes 160x160 size RGB color images as input. The images are aligned, cropped and resized using AlignDlib\footnote{\url{https://github.com/krasserm/face-recognition}} to meet the required size. The probe data with labelled quality score is divided into 70-30 training and test data split. The loss function criterion used for the task is log mean squared error. For optimization, Stochastic Gradient Descent (SGD) is used with the learning rate of 0.001, momentum of 0.99, and weight decay of $10^{-5}$. A mini-batch size of 64 is used and the training is iterated over 30 epochs. The experiment was done using NVIDIA GTX 1080 GPU with 8GB memory.

\begin{equation}
   Loss = \sqrt{\frac{1}{N}\sum_{i=1}^{N}\left ( \log \left ( y_i \right ) -  \log \left ( \hat{y}_i \right ) \right )}
\end{equation}

\subsection{Results}

The model is evaluated on Georgia Tech Face Database containing 50 subjects with 15 images each. Fig \ref{tsne} shows 2-dimensional t-Distributed Stochastic Gradient Embedding (t-SNE) visualization of the 128-dimensional FaceNet embeddings for seven subjects. Each colour label corresponds to a subject and it can be seen that facial images of same subjects are well clustered together and different clusters are well separable and distinguisable. The metrics used to evaluate the model are False Acceptance Rate (FAR) and False Rejection Rate (FRR).

The set of false accepts (FA) or false positives can be defined as
\begin{equation} \label{far}
FA(d) = \{ (i,j) \in P_{diff}, D(x,y) \leq d \}
\end{equation}

The set of true rejects (TR) or true negative can be defined as
\begin{equation} \label{frr}
TR(d) = \{ (i,j) \in P_{same}, D(x,y) > d \}
\end{equation}

\begin{figure}[h]
\centering
        \begin{subfigure}[b]{0.5\textwidth}
                \centering
                \includegraphics[width=\linewidth]{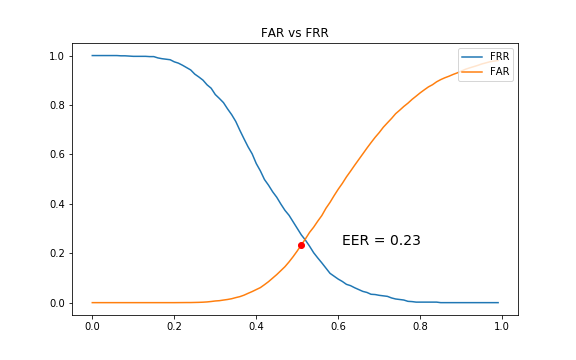}
                \caption{Plot of FAR and FAR}
                \label{fig:eer}
        \end{subfigure} \hfill
        \begin{subfigure}[b]{0.49\textwidth}
                \centering
                \includegraphics[width=\linewidth]{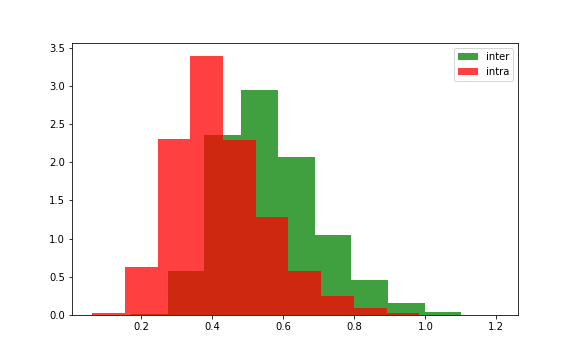}
                \caption{Inter and Intra class distribution}
                \label{fig:dist}
        \end{subfigure}\hfill
        
        \caption{Some statistics}\label{fig:animals}
\end{figure}

Therefore, False Acceptance Rate (FAR) and False Rejection Rate (FRR) is defined as
\begin{equation}
    FAR(d) = \frac{FA(d)}{|P_{diff}|}, \;\;\;\;\; FRR(d) = \frac{TR(d)}{|P_{same}|} 
\end{equation}
The predicted score for different poses of a subject is shown in Fig. \ref{score}. The metric False Acceptance Rate and False Rejection Rate was calculated and shown in Fig. \ref{fig:eer}. As seen from the plot, an Equal Error Rate of 23\% is achieved on the Georgia Tech Face Database.

\begin{figure}[!h]
\centering
\includegraphics[width=\linewidth]{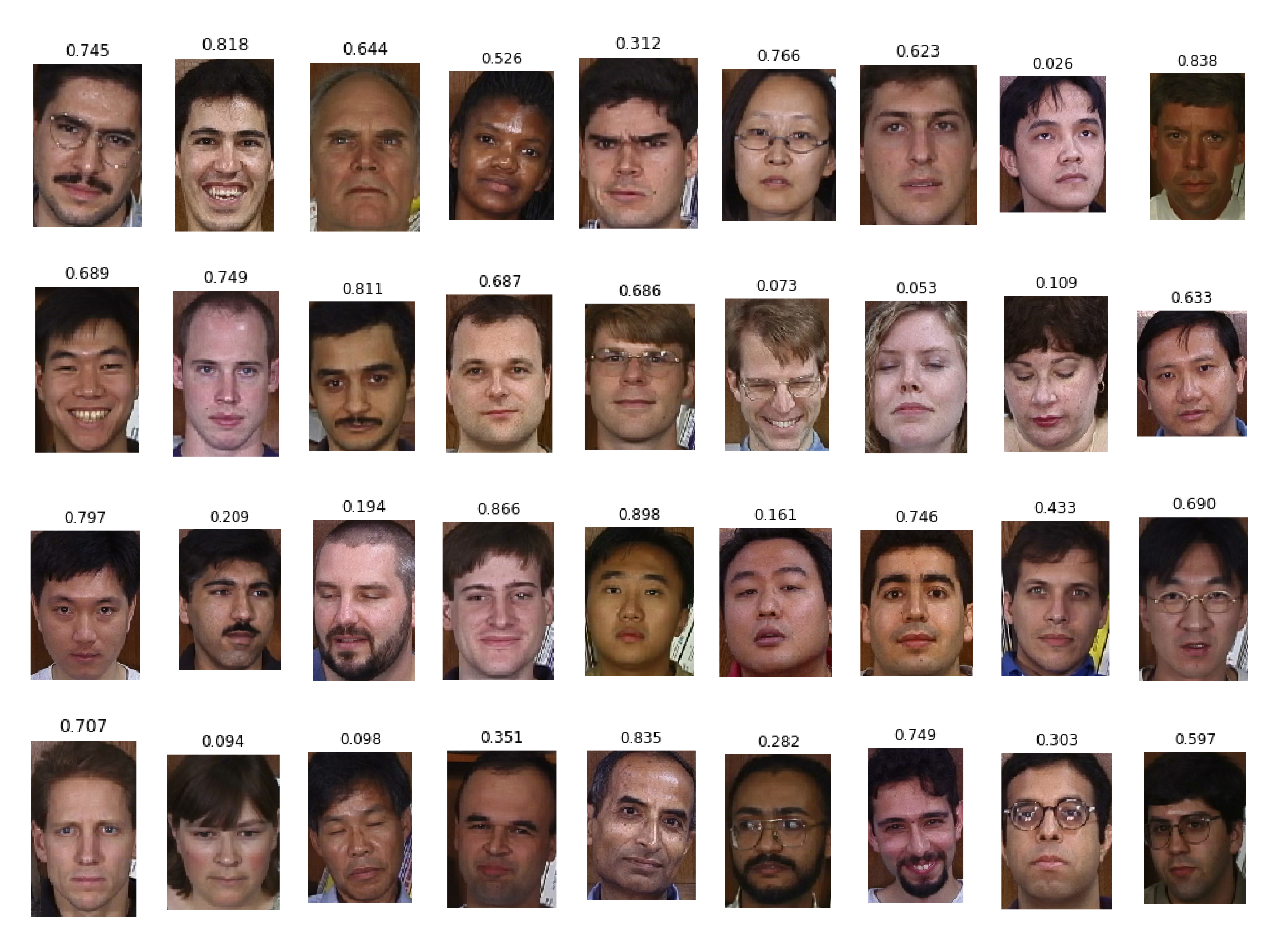}
\caption{Score for few subjects}
\label{all}
\end{figure}

\section{Conclusion}

Face image quality assessment can be used as a pre-processing step in any facial recognition system. This type of assessment is crucial in Negative Identification Systems such as banks, airports, etc. where criminals try to escape their recognition and present low quality image. By simple thresholding of the quality score, face verification can be restricted. In this work, a simple end-to-end approach has been proposed to evaluate face image quality presented to a facial recognition system. The model used is based on Inception v3 architecture and is trained in a supervised way using the quality score labels generated. Also, the pre-trained weights from FaceNet has been used as an initialization for the model to achieve better performance. Since the face quality score labels are not readily available, these labels were generated as normalized similarity score by comparison of face images using FaceNet embeddings. 

The model is evaluated using the metric False Acceptance Rate (FAR) and False Rejection Rate (FRR), the plot for which is shown in Fig. \ref{fig:eer}. In spite of training with such low volume of data, the model performs pretty well on the task. An Equal Error Rate of 23\% is achieved on the Georgia Tech Face Database which contains images of people from multiple ethnic groups and backgrounds. Deep learning models are data hungry and require to be trained with big datasets to generalize better. If we train the model with more volume of data, we can achieve better performance and lower EER.



\bibliographystyle{plain} 

\bibliography{bibliography}

\begin{thebibliography}{10}

\bibitem{gtdb}
Georgia tech face database,
  \url{http://www.anefian.com/research/face_reco.htm}, 1999.

\bibitem{rowden}
Lacey Best-Rowden and Anil~K. Jain.
\newblock Automatic face image quality prediction.
\newblock {\em CoRR}, abs/1706.09887, 2017.

\bibitem{int2}
Haoqiang Fan, Zhimin Cao, Yuning Jiang, Qi~Yin, and Chinchilla Doudou.
\newblock Learning deep face representation.
\newblock {\em CoRR}, abs/1403.2802, 2014.

\bibitem{meth1}
P.~Grother and E.~Tabassi.
\newblock Performance of biometric quality measures.
\newblock {\em IEEE Transactions on Pattern Analysis and Machine Intelligence},
  29(4):531--543, April 2007.

\bibitem{int1}
Patrick~J Grother, George~W Quinn, and P~Jonathon Phillips.
\newblock Report on the evaluation of 2d still-image face recognition
  algorithms.
\newblock {\em NIST interagency report}, 7709:106, 2010.

\bibitem{lfw}
Gary~B Huang, Marwan Mattar, Tamara Berg, and Eric Learned-Miller.
\newblock Labeled faces in the wild: A database forstudying face recognition in
  unconstrained environments.
\newblock In {\em Workshop on faces in'Real-Life'Images: detection, alignment,
  and recognition}, 2008.

\bibitem{facenet}
Florian Schroff, Dmitry Kalenichenko, and James Philbin.
\newblock Facenet: A unified embedding for face recognition and clustering.
\newblock In {\em The IEEE Conference on Computer Vision and Pattern
  Recognition (CVPR)}, June 2015.

\bibitem{work2}
H.~Sellahewa and S.~A. Jassim.
\newblock Image-quality-based adaptive face recognition.
\newblock {\em IEEE Transactions on Instrumentation and Measurement},
  59(4):805--813, April 2010.

\bibitem{meth3}
E~Tabassi and C.L. Wilson.
\newblock A novel approach to fingerprint image quality.
\newblock 2:II -- 37, 10 2005.

\bibitem{meth2}
P.~Viola and M.~Jones.
\newblock Robust real-time face detection.
\newblock In {\em Proceedings Eighth IEEE International Conference on Computer
  Vision. ICCV 2001}, volume~2, pages 747--747, July 2001.

\bibitem{work3}
Yongkang Wong, Shaokang Chen, Sandra Mau, Conrad Sanderson, and Brian~C Lovell.
\newblock Patch-based probabilistic image quality assessment for face selection
  and improved video-based face recognition.
\newblock In {\em Computer Vision and Pattern Recognition Workshops (CVPRW),
  2011 IEEE Computer Society Conference on}, pages 74--81. IEEE, 2011.

\bibitem{work1}
Zhiguang Yang, Haizhou Ai, Bo~Wu, Shihong Lao, and Lianhong Cai.
\newblock Face pose estimation and its application in video shot selection.
\newblock 1:322-- 325 Vol.1, 09 2004.

\bibitem{tf}
Jason Yosinski, Jeff Clune, Yoshua Bengio, and Hod Lipson.
\newblock How transferable are features in deep neural networks?
\newblock {\em CoRR}, abs/1411.1792, 2014.

\bibitem{lit}
W.~Zhao, Rama Chellapa, P.J. Phillips, and A.~Rosenfeld.
\newblock Face recognition: A literature survey.
\newblock {\em ACM Comput. Surv.}, 35:399 -- 458, 2003.

\end{thebibliography}

\end{document}